\documentclass{article}

\usepackage{subfiles}
\usepackage[pdftex]{graphicx}
\usepackage{amsthm}
\usepackage{mathtools}
\usepackage{booktabs}
\usepackage{multirow}
\usepackage{siunitx}
\usepackage{wrapfig}

\usepackage{arxiv}
\usepackage[utf8]{inputenc} 
\usepackage[T1]{fontenc}    
\usepackage{hyperref}       
\usepackage{url}            
\usepackage{booktabs}       
\usepackage{amsfonts}       
\usepackage{nicefrac}       
\usepackage{microtype}      
\usepackage{cleveref}       
\usepackage{lipsum}         
\usepackage{graphicx}
\usepackage{natbib}
\usepackage{doi}

\title{Maximum Entropy On-Policy Actor-Critic via Entropy Advantage Estimation}

\newif\ifuniqueAffiliation
\uniqueAffiliationtrue

\author{
    Jean Seong Bjorn Choe \\
    School of Electrical Engineering \\
    Korea University \\
    \texttt{garangg@korea.ac.kr} \\
    \And
    Jong-Kook Kim \\
    School of Electrical Engineering \\
    Korea University \\
    \texttt{jongkook@korea.ac.kr}
}

\hypersetup{
pdftitle={Maximum Entropy On-Policy Actor-Critic via Entropy Advantage Estimation},
pdfsubject={cs.AI},
pdfauthor={J. S. Bjorn Choe, Jong-Kook Kim},
pdfkeywords={MaxEnt RL, Reinforcement Learning, Policy Optimization},
}

\newtheorem{theorem}{Theorem}

\newcommand{\expect}[2]{\mathbb{E}_{#1}\left[{#2}\right]}
\newcommand{\avg}[1]{\hat{\mathbb{E}}_t{\left[{#1}\right]}}
\newcommand{\ent}{\mathcal{H}}

\newcommand{\vent}{V^\pi_{\ent}}
\newcommand{\qent}{Q^\pi_\ent}
\newcommand{\aent}{A^\pi_\ent}
\newcommand{\gent}{\gamma_\ent}
\newcommand{\lent}{\lambda_\ent}

\newcommand{\asoft}{\tilde{A}^\pi}
\newcommand{\asoftt}{\tilde{A}^\pi_t}

\newcommand{\lpt}{\log\pi(a_t|s_t)}

\begin{document}
\maketitle

\begin{abstract}
Entropy Regularisation is a widely adopted technique that enhances policy optimisation performance and stability.
A notable form of entropy regularisation is augmenting the objective with an entropy term, thereby simultaneously optimising the expected return and the entropy. This framework, known as maximum entropy reinforcement learning (MaxEnt RL), has shown theoretical and empirical successes. However, its practical application in straightforward on-policy actor-critic settings remains surprisingly underexplored. We hypothesise that this is due to the difficulty of managing the entropy reward in practice. This paper proposes a simple method of separating the entropy objective from the MaxEnt RL objective, which facilitates the implementation of MaxEnt RL in on-policy settings. Our empirical evaluations demonstrate that extending Proximal Policy Optimisation (PPO) and Trust Region Policy Optimisation (TRPO) within the MaxEnt framework improves policy optimisation performance in both MuJoCo and Procgen tasks. Additionally, our results highlight MaxEnt RL's capacity to enhance generalisation. 
\end{abstract}

\section{Introduction}
Entropy regularisation is pivotal to many practical deep reinforcement learning (RL) algorithms. Practical algorithms such as Trust Region Policy Optimization (TRPO) \citep{schulman2015trust} penalise the policy improvement or greedy step using Kullback-Leibler (KL) divergence (also called as relative entropy) to regularise the deviations between consecutive policies. This method, often termed KL regularisation, has been the foundational approach for contemporary deep RL algorithms \citep{vieillard2020leverage,geist2019theory}. 

Another critical approach is to regularise the policy evaluation step by augmenting the conventional RL task objective with an entropy term, thereby directing policies toward areas of higher expected trajectory entropy. This scheme is often called Maximum Entropy RL (MaxEnt RL) \citep{ziebart2010modeling,haarnoja2018soft,levine2018reinforcement}. MaxEnt RL formulation is known to improve the exploration and robustness of policies by promoting stochasticity \citep{eysenbach2019if,eysenbach2021maximum}. In practice, MaxEnt RL can simply be implemented by adding an entropy reward to the original task reward.

Recent theoretical advancements inspired by the Mirror Descent theory have developed a unified view of these approaches \citep{vieillard2020leverage,geist2019theory,tomar2020mirror}, suggesting that their combination could lead to faster convergence to the solution of the regularised objective \citep{shani2020adaptive}. Furthermore, the latest studies on policy gradient (PG) methods have shown the effectiveness of the MaxEnt RL in accelerating the convergence of PG algorithms \citep{mei2020global,agarwal2021theory,cen2022fast}. However, despite the enticing theoretical support, its practical application remains underexplored, particularly in stochastic policy gradient methods in on-policy settings.

    \begin{wrapfigure}{R}{0.5\textwidth}
        \includegraphics[width=0.45\textwidth]{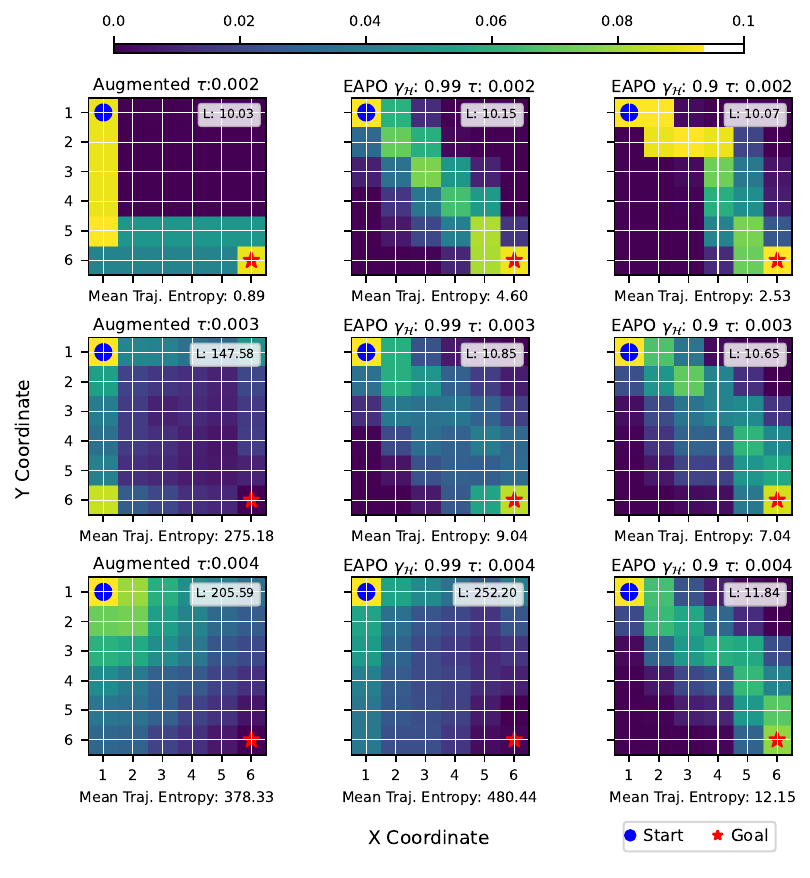}
        \caption{
            The normalised state visitation counts from 100 rollouts with policies trained on the modified \texttt{MiniGrid-Empty-8x8} task using a naive MaxEnt algorithm (PPO with the augmented entropy reward) and EAPO using 2 different discount factors $\gent \in (0.9, 0.99)$ and TD$(0)$ for entropy estimation. We compare 3 different temperatures $\tau\in(0.002,0.003,0.004)$.
            A discount factor $\gamma_V=0.99$ is used for the task reward. $L$ is the mean length of trajectories, with agents aiming to minimise it (10 is optimal).
            See Appendix \ref{app:example-empty} for more details.
        }
        \label{fig:minigrid-empty}
    \end{wrapfigure}


We hypothesise that this research gap is potentially attributed to the difficulty of handling the entropy reward in practice. \citet{yu2022you} empirically analysed the problematic nature of the entropy reward using Soft Actor-Critic (SAC) \citep{haarnoja2018soft}, an off-policy MaxEnt algorithm. Authors pointed out that in an episodic setting, the entropy return is largely correlated to the episode's length, thereby rendering the policy overly optimistic or pessimistic, and even in infinite-horizon settings, the entropy reward can still obscure the task reward. 

Inspired by this observation, we proposed a simple but practical approach to control the impact of the entropy reward. In this paper, we introduce Entropy Advantage Policy Optimisation (EAPO), a method that estimates the task and entropy objectives of the regularised (soft) objective separately. By employing a dedicated discount factor for the entropy reward and utilising Generalised Advantage Estimation (GAE) \citep{schulman2015high} on each objective separately, EAPO controls the effective horizon of the entropy return estimation and the entropy regularisation on policy evaluation. EAPO's simplicity requires only minor modifications to existing advantage actor-critic algorithms. This work extends the well-established PPO \citep{schulman2017proximal} and TRPO \citep{schulman2015trust}.

Figure \ref{fig:minigrid-empty} illustrates the challenge of learning the MaxEnt policy for an episodic task using a naive implementation that simply augments the task reward with an entropy reward. In this task, the agent is required to reach the goal state while performing the minimum number of actions. The naive MaxEnt agent fails to learn the optimal stochastic policy, resulting in two failure modes: acting almost deterministically when the temperature $\tau$ is low or wandering around indefinitely when $\tau$ is high. In contrast, EAPO successfully achieves the near-optimal stochastic policy by utilising TD$(0)$ learning \citep{sutton2018reinforcement} (i.e., set GAE $\lambda$ to $0$) for the entropy objective. Additionally, the example demonstrates that lowering the discount factor for the entropy estimation $\gent$ helps prevent the inflation of the entropy reward \citep{yu2022you} and reduces sensitivity to the temperature.

In this work, we empirically demonstrate that EAPO allows the development of a practical MaxEnt on-policy actor-critic algorithm. We test the efficacy of EAPO within deterministic environments with the discrete action space to align with existing theories on the MaxEnt formulation \citep{levine2018reinforcement} and the softmax policy gradient methods \citep{mei2020global}. Specifically, we evaluate the general training performance on 4 discretised \citep{tang2020discretizing} Mujoco continuous control tasks, including those with an infinite horizon \citep{todorov2012mujoco}. We also test the robustness in 16 Procgen episodic environments \citep{cobbe2020leveraging}. Additionally, we examine the usefulness of a MaxEnt policy in MiniGrid DoorKey environment  \citep{MinigridMiniworld23}.

\section{Background}
\subsection{Preliminaries}
This work considers a finite discounted deterministic Markov Decision Process (MDP) $\langle\mathcal{S}, \mathcal{A}, r, \rho, \mathcal{T}, \gamma_V, \gamma_\ent \rangle$, where $\mathcal{S}$ is the set of states $s$ and 
$\mathcal{A}$ is the set of actions $a$, and $\rho$ is the initial state distribution. $\mathcal{T}$ is the deterministic transition function $\mathcal{T}: \mathcal{S}\times\mathcal{A}\mapsto \mathcal{S}$, and $r$ is the reward function $r: \mathcal{S}\times\mathcal{A}\mapsto\mathbb{R}$. $\gamma_V$ and $\gamma_\ent$ are the discount factors. 

We define the value function of state $s$ under the policy $\pi$ as 
\begin{equation*}
    V^\pi(s)\coloneqq \expect{
            s_0=s,  a_t\sim\pi(\cdot|s_t), \\ s_{t+1}=\mathcal{T}(s_t,a_t)
        }{\sum_{t=0}^{\infty}\gamma_V^{t}r(s_t,a_t)}.
\end{equation*}

Also, the action-value of performing action $a$ at state $s$ under the policy $\pi$  is
\begin{equation*}
    Q^\pi(s,a) \coloneqq \expect{
            s_0=s, a_0=a, \\ a_{t>0}\sim\pi(\cdot|s_t), \\ s_{t+1}=\mathcal{T}(s_t,a_t)
        }{ \sum_{t=0}^{\infty}\gamma_V^{t}r(s_t,a_t)}.
\end{equation*}
And define the advantage function $A^\pi$ as
$
    \label{eq:adv_f}
    A^\pi(s,a) \coloneqq Q^\pi(s,a) - \expect{\pi(\cdot|s)}{Q^\pi(s,\cdot)} 
    = Q^\pi(s,a) - V^\pi(s)
$.
We also define the discounted policy-induced trajectory entropy, or the entropy rate of state $s$ under policy $\pi$ as
\begin{equation}\label{eq:hfunc}
    \vent(s)\coloneqq \expect{
        \substack{s_0=s,  a_t\sim\pi(\cdot|s_t), \\ s_{t+1}=\mathcal{T}(s_t,a_t)}}
        {
            \sum_{t=0}^{\infty}-\gamma_\ent^{t}\lpt
        }.
\end{equation}
This trajectory entropy represents the Shannon entropy of the possible future trajectories' distribution in an MDP with deterministic dynamics \citep{levine2018reinforcement,tiapkin2023fast}. The objective of Maximum Entropy Reinforcement Learning (MaxEnt RL), or often Regularised MDPs \citep{geist2019theory,neu2017unified} is to maximise the expectation of the sum of the value and the trajectory entropy with respect to the initial state distribution:

\begin{align}\label{eq:soft_obj}
    J(\pi) &= \expect{
    \substack{s_0\sim \rho,  a_t\sim\pi, \\ s_{t+1}=\mathcal{T}(s_t,a_t)}}
    {
        \sum_{t=0}^{\infty}\gamma_V^{t}
        r(s_t,a_t)-\gamma_\ent^t\tau\lpt
    } \\
    &= \expect{s_0\sim \rho}{V^\pi(s_0) + \tau \vent(s_0)}
\end{align}
where the temperature parameter $\tau\geq0$ is a hyperparameter to be controlled to balance the significance between these two objectives, and we introduce the distinct discount factors.

\subsection{Soft advantage function}
Analogous to the definition of the action-value function $Q^\pi$ as the expected cumulative rewards after selecting an action $a$ \citep{sutton2018reinforcement}, we define $Q^\pi_\ent$ as the expected future trajectory entropy after selecting an action:
\begin{align} \label{eq:qdef}
    Q^\pi_\ent(s_t,a_t) &\coloneqq \gent\vent(\mathcal{T}(s_t,a_t)) = \gent\vent(s_{t+1}).
\end{align}
The definition arises naturally from the consideration that uncertainty exists due to the stochastic policy at the current state, which has settled by the time an action is performed. Consequently, the $\qent$ is simply the discounted trajectory entropy of the next state determined by the deterministic transition function.

From the recursive relation of trajectory entropy from (\ref{eq:hfunc}) and the defintion (\ref{eq:qdef}), the following relation is derived:
\begin{equation}\label{eq:hqh}
    \vent(s_t) = \expect{a_t\sim\pi(\cdot|s_t)}{-\lpt  + \qent(s_t,a_t)}.
\end{equation}
We now define the entropy advantage function $A^\pi_\ent$ analogous to (\ref{eq:adv_f}):
\begin{align}\label{eq:ea}
    \aent(s_t,a_t) &\coloneqq \qent(s_t,a_t) - \expect{a\sim\pi(\cdot|s_t)}{\qent(s_t,a)} \nonumber\\
                    &= \qent(s_t,a_t) - \vent(s_t) + \expect{a\sim\pi(\cdot|s_t)}{-\log\pi(a|s_t)}.
\end{align}

We let $\tilde{V}^\pi(s)\coloneqq V^\pi(s)+\tau\vent(s)$ as the soft value function, and let $\tilde{Q}^\pi(s,a)\coloneqq Q^\pi(s)+\tau Q^\pi_\ent(s,a)$ as the soft Q-function. Finally, we define the soft advantage function:
\begin{align}\label{eq:sa}
    \asoft(s_t,a_t) &\coloneqq A^\pi(s_t,a_t) + \tau A^\pi_\ent(s_t,a_t) \\
        &= Q^\pi(s_t,a_t) - V^\pi(s_t) + \tau (\qent(s_t,a_t) - \vent(s_t) + \expect{a\sim\pi(\cdot|s_t)}{-\log\pi(a|s_t)}) \notag \\
        &= \tilde{Q}^\pi(s_t, a_t) - \tilde{V}^\pi(s_t)  + \tau \expect{a\sim\pi(\cdot|s_t)}{-\log\pi(a|s_t)}.
\end{align}

\subsection{Soft policy gradient theorem}
\citet{shi2019soft} showed that it is possible to optimise the soft objective using direct policy gradient from samples. Thus, we can use the soft advantage function to find the policy that maximises the MaxEnt RL objective.
\begin{theorem}[Soft Policy Gradient]
    Let  $J(\pi)$ the MaxEnt RL objective defined in \ref{eq:soft_obj}. And $\pi_\theta(a|s)$ be a parameterised policy. Then,
    \begin{equation}
        \nabla_\theta J(\pi_\theta) = \expect{\substack{s_0\sim \rho, \\ a_t\sim\pi, \\ s_{t+1}= \mathcal{T}(s_t,a_t)} }{(\gamma_V^t A(s_t,a_t)+\gent^t \tau\aent(s_t,a_t))\nabla_\theta\log\pi_\theta(a_t|s_t)}.
    \end{equation}
\end{theorem}

We provide the proof in Appendix \ref{app:spgt}. While the exact soft policy gradient theorem requires the corresponding exponential discount term for each advantage estimate, we use the approximate policy gradient in this work:
\begin{equation}
    \nabla_\theta J(\pi_\theta) \approx \expect{\substack{s_0\sim \rho, \\ a_t\sim\pi, \\ s_{t+1}= \mathcal{T}(s_t,a_t)} }{\asoft(s_t,a_t)\nabla_\theta\log\pi_\theta(a_t|s_t)}.
\end{equation}
It is worth noting that when the exact gradient is known, \citet{mei2020global} proved that the soft policy gradient has the global convergence property and may converge faster than the policy gradient without entropy regularisation despite the objective being biased. However, in our practical setup, this is not guaranteed.

\section{Related works}
One of the most prominent aspects of the MaxEnt RL formulation that has been studied is the ability to connect policy gradient methods and off-policy value-based methods that learn the soft $Q$-function \citep{haarnoja2018soft,nachum2017bridging,o2016combining,schulman2017equivalence}. However, this work focuses on the soft policy gradient using the soft advantage estimation in an on-policy setting.

\citet{shi2019soft} explored the soft policy gradient method, emphasising its inherent simplicity. While they employed the soft $Q$-function to guide the policy gradient, linking their method to off-policy techniques, EAPO leverages the variance-reduced estimation of the entropy advantage function to formulate the soft advantage function suitable for on-policy algorithms. Moreover, \citet{shi2019soft} introduced additional techniques to mitigate the challenging task of estimating the soft $Q$-function. However, EAPO can seamlessly integrate with existing techniques, such as value function normalisation, and GAE, due to its structural equivalence between its method for estimating the entropy advantage function and the conventional advantage function.

A more common approach to applying entropy regularisation to PG methods is to add an entropy cost term to the sample-based policy gradient estimator to maximise the policy entropy at each sampled state, retaining the stochasticity of the policy during optimisation process \citep{mnih2016asynchronous,schulman2017proximal}. While the entropy bonus term seeks to maximise policy entropy at visited states, MaxEnt RL directs a policy toward regions of higher expected trajectory entropy, albeit at the cost of bias imposed on the objective \citep{levine2018reinforcement,schulman2017equivalence}.  Despite its empirical success, this method remains a heuristic approach without solid theoretical understanding\citep{ahmed2019understanding}. In this work, we show that the use of MaxEnt RL can replace the traditional entropy cost term.

The algorithms studied in this work can be seen as an instance of algorithm that both KL regularisation and entropy regularisation are applied \citep{vieillard2020leverage,geist2019theory}. However, this work does not analyse how these two types of regularisations interact.

 \citet{yu2022you} studied the harmful effect of having entropy regularisation on the policy evaluation step and has motivated this research. Unlike their work, we draw a positive conclusion for the use of entropy rewards. In our experiments comparing the entropy cost method, using the MaxEnt framework instead of the entropy cost consistently improves the performance of on-policy algorithms.

 Recent studies have investigated the theoretical properties of policy gradient methods \citep{agarwal2021theory,mei2020global,cen2022fast}. Notably, the authors have shown that combining Natural Policy Gradient (NPG) methods \citep{kakade2001natural} and the entropy-regularised MDPs can speed up the convergence. Although some authors \citep{khodadadian2021linear,shani2020adaptive} have drawn the connection between PPO and NPG methods rigorously in theoretical settings, it remains unclear whether PPO-based EAPO, which extends PPO to the Regularised MDP setting, can benefit from the theoretical guarantees.

\section{Proposed method}
\subsection{Overview}
In this section, we develop our Entropy Advantage Policy Optimisation (EAPO) method. At its core, EAPO independently estimates both the value advantage function and the entropy advantage function and combines them to derive the soft advantage function.
EAPO adopts a separate prediction head to the conventional value critic to approximate the trajectory entropy of a state, which is then used for entropy advantage estimation.
We extend the PPO \citep{schulman2017proximal} and TRPO \citep{schulman2015trust} by substituting the advantage estimate with the soft advantage estimate and omitting the entropy bonus term.

\subsection{Entropy advantage estimation}
The entropy advantage $\aent$ is estimated from the sampled log probabilities of the behaviour policy. We utilise the Generalised Advantage Estimation (GAE) \citep{schulman2015high} for a variance-reduced estimation of the entropy advantage:
\begin{align}
  \hat{A}^{\ent, \text{GAE}(\lambda_\ent,\gamma_\ent)}(s_t, a_t) \coloneqq \sum^\infty_{l=0}(\lambda_\ent\gamma_\ent)^l\delta^\ent_{t+l},
\end{align}
where $\delta^\ent_{t}\coloneqq -\log\pi(a_t|s_t) + \gamma_\ent\vent(s_{t+1}) - \vent(s_t)$, and $\gamma_\ent$ and $\lambda_\ent$ are the discount factor and GAE lambda for entropy advantage estimation, respectively.
Note that the equation is the same as the GAE for the conventional advantage, except the reward term is replaced by the negative log probability.
This simplicity is also consistent with the remark that the only modification required for the MaxEnt policy gradient is to add the negative log probability term to the reward at each time step \citep{levine2018reinforcement}. 

\subsection{Entropy critic}
An entropy critic network, parameterised by $\omega$, approximates the trajectory entropy $\vent$, and it is trained by minimising the mean squared error
$
  L^\ent(\omega) \coloneqq \avg{\frac{1}{2}\left(\vent(s_t;\omega)  - \hat{V}^\pi_\ent(s_t)\right)^2},
$
where the trajectory entropy estimate $\hat{V}^\pi_\ent(s_t)$ is calculated using TD$(\lent)$: $\hat{V}^\pi_\ent(s_t) = \hat{A}^{\ent, \text{GAE}(\lambda_\ent,\gamma_\ent)}(s_t,a_t) + \vent(s_t;\omega)$. Throughout the conducted experiments, we implemented the entropy critic network to share its parameters with the return value critic $V^V_\phi$, with only the final linear layers for outputting its prediction distinct. This form of parameter sharing allows minimal computational overhead to implement EAPO.

Further, we employ the PopArt normalisation \citep{van2016learning} to address the scale difference of entropy and return estimates. It is important to note that the negative log probability $-\log\pi(a_t|s_t)$ is collected for every timestep. In contrast, the reward can be sparse, leading to significant magnitude variations based on the dynamics of the environment \citep{hessel2019multi}. This discrepancy can pose challenges, especially when using a shared architecture. Thus, utilising the value normalisation technique like PopArt is pivotal for the practical implementation of EAPO.

\subsection{Entropy advantage policy optimisation}
Subsequently, we integrate the entropy advantage with the standard advantage estimate $\hat{A}^\pi_V$, also computed using GAE and return value critic parameterised by $\phi$, analogously to the entropy advantage estimation process we describe above. Then the soft advantage function $\asoft$ is
\begin{align}
  \asoft(s_t,a_t) = \hat{A}^{V, \text{GAE}(\lambda_V,\gamma_V)}(s_t, a_t) + \tau  \hat{A}^{\ent, \text{GAE}(\lambda_\ent,\gamma_\ent)}(s_t,a_t),
\end{align}
where $\hat{A}^{V, \text{GAE}(\lambda_V,\gamma_V)}$ is the value advantage estimation using GAE.
Finally, we substitute the estimated conventional advantage function in the policy objective of PPO and TRPO with $\asoft$. The PPO objective function becomes:

\begin{align}
  L(\theta, \phi, \omega) &= \avg{\min(r_t(\theta)\asoft(s_t,a_t), \text{clip}(r_t(\theta),1-\epsilon,1+\epsilon)\asoft(s_t,a_t))} \nonumber \\ 
  &+ c_1(L^V(\phi) + c_2 L^\ent(\omega)),
\end{align}
where $r_t(\theta)$ is the probability ratio between the behaviour policy $\pi_{\theta_\text{old}}(a_t|s_t)$ and the current policy $\pi_\theta(a_t|s_t)$, and $c_1$, $c_2$ and $\epsilon$ are 
hyperparameters to be adjusted. The value critic loss $L^V$ is also defined by the mean square error, $L^V(\phi)=\avg{\frac{1}{2}\left(V(s_t;\phi)  - \hat{V}^\pi(s_t)\right)^2}$ where $\hat{V}^\pi$ is the return value estimate.

Similarly, the optimisation problem of TPRO becomes:
\begin{align}
  \max_{\theta\in\Theta}\avg{r_t(\theta)\asoftt(s,a)},\;\;s.t. \;\;
  \avg{\text{KL}(\pi_{\theta_\text{old}} || \pi_\theta)} \leq \delta,
\end{align}
where $\delta$ is a hyperparameter.

\section{Experiments}
In this section, we evaluate the policy optimisation performance of EAPO against the corresponding baseline on-policy algorithms, PPO and TRPO. Specifically, we assess the optimisation efficiency for episodic tasks and the generalisation capability of EAPO on 16 Procgen \citep{cobbe2020leveraging} benchmark environments. Moreover, we investigate EAPO's efficacy on continuing control tasks using 4 discretised popular MuJoCo \citep{todorov2012mujoco} environments, and we analyse the impact of hyperparameters $\tau$, $\gent$ and $\lent$. Finally, we include \texttt{MiniGrid-DoorKey-8x8} environment \citep{MinigridMiniworld23} to examine if EAPO can help solve the hard exploration task.

We implemented EAPO using Stable-baseline3 \citep{raffin2019stable} and conducted experiments on environments provided by the Envpool \citep{weng2022envpool} library.
All empirical results are averaged over from 10 random seeds, with a 95\% confidence interval indicated.

For the hyperparameter selection, we conducted a brief search for baseline algorithm hyperparameters that perform reasonably well, tuning only the EAPO-specific hyperparameters such as $\gent$ to ensure fair comparisons. Implementation details and hyperparameters are reported in Appendix \ref{app:hps}.

\subsection{Procgen benchmark environments}
We evaluate the performance and the generalisation capability of PPO-based EAPO on the 16 Procgen benchmark environments \citep{cobbe2020leveraging}. These environments have a discrete action space and use raw images as observations. Procgen suite includes episodic tasks with both positive (e.g., BigFish) and negative (e.g., Climber) correlations between the episode length and the return, making them suitable for testing MaxEnt RL algorithms. Following the evaluation procedure in \citep{cobbe2020leveraging}, we trained agents on 200 procedurally generated levels and test the performance as the mean episodic return on 100 unseen levels for each environment. We use the \textit{easy} difficulty setting. We employed a single set of hyperparameters for all environments. 
For hyperparameter tuning, we used three representative environments: BigFish, Climber and Dodgeball. Each represents different levels of correlation between episode length and the return. BigFish requires the agent to survive as long as possible, Climber allows indefinite exploration, and in Dodgeball, the agent must survive during the first phase and reach the goal quickly in the second stage.

\begin{figure*}
    \begin{minipage}[t]{0.48\textwidth}
        \centering
        \includegraphics[width=0.99\textwidth]{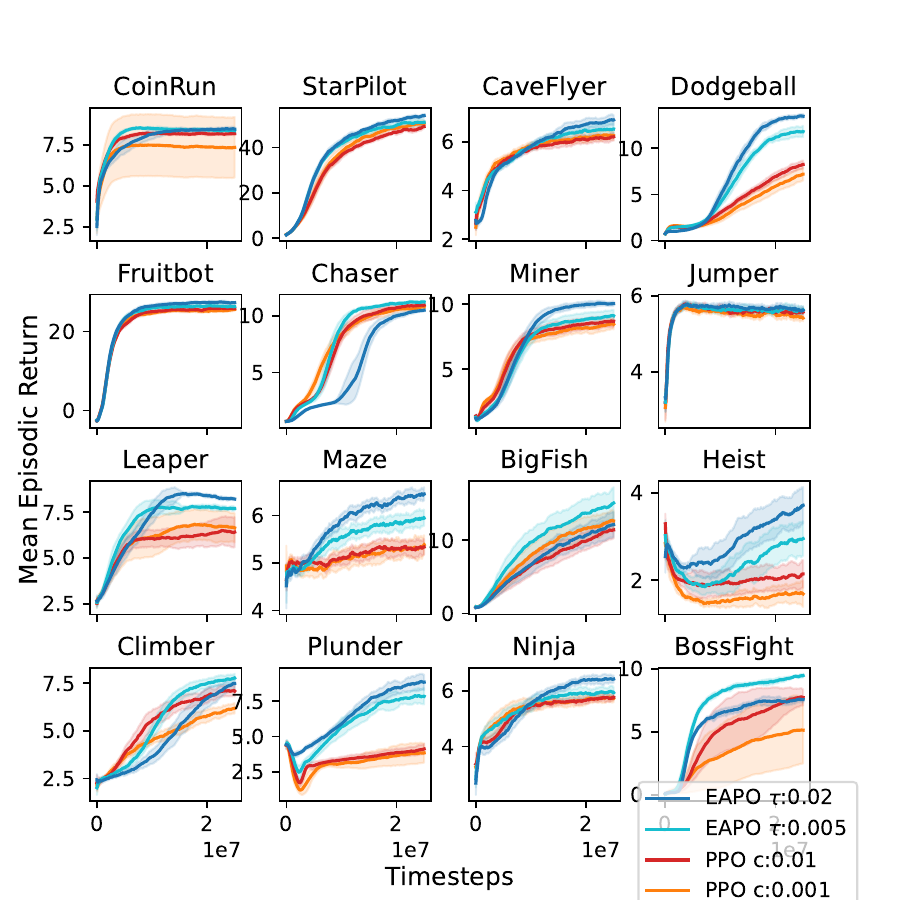}
    \end{minipage}
    \hfill
    \begin{minipage}[t]{0.5\textwidth}
        \centering
        \includegraphics[width=0.99\textwidth]{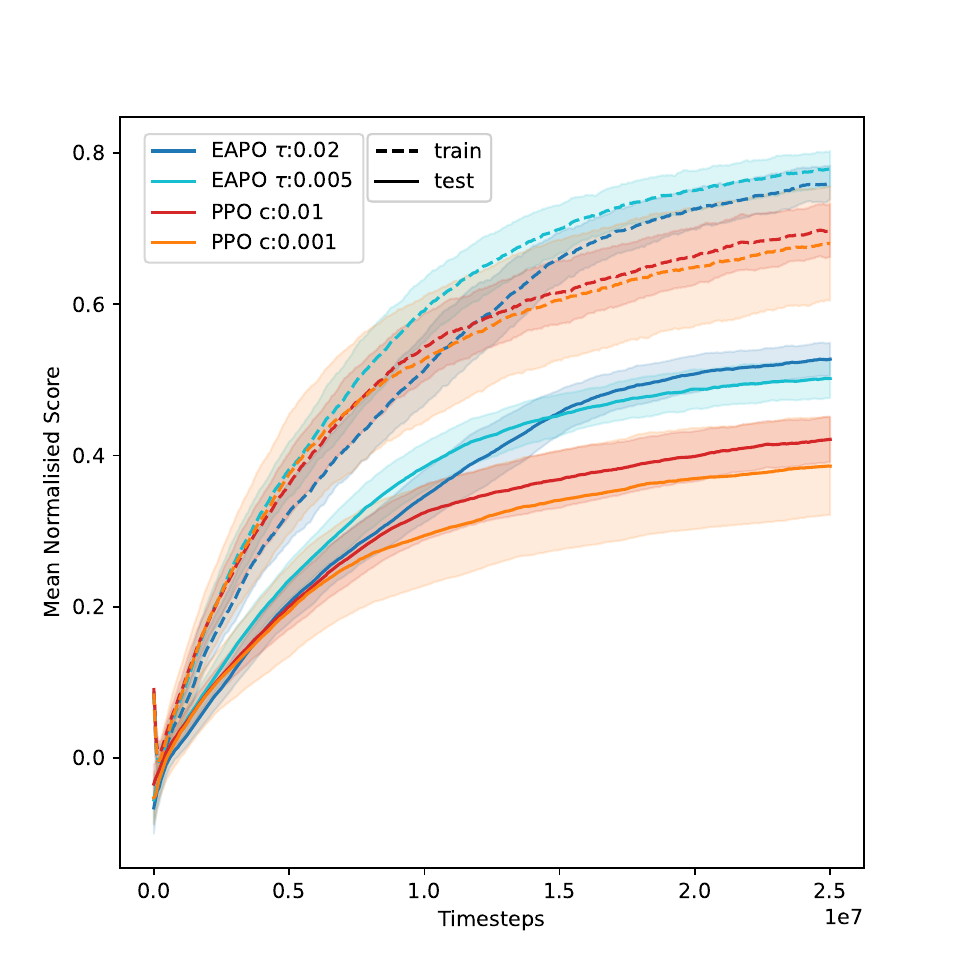} 
    \end{minipage}
    \caption{
        \textbf{Left}: Generalisation test results of EAPO agents with $\gent=0.8$, $\lent=0.95$ , and two different temperatures $\tau=0.02$ and $\tau=0.005$ against PPO agents with entropy coefficients of $0.001$ and $0.01$ on 16 Procgen \citep{cobbe2020leveraging} benchmark environments. Agents are evaluated on 100 levels unseen during the training.  EAPO consistently outperforms or at least matches PPO in all environments. Results are averaged over 10 seeds, and the shaded area indicates the 95\% confidence interval. \textbf{Right}: The mean normalised score for both test and training, computed according to \citep{cobbe2020leveraging}.
    }
    \label{fig:procgen}
\end{figure*}

Figure \ref{fig:procgen} and Table \ref{table:procgen} summarise the generalisation test results of EAPO and baseline PPO agents with varying $\tau$ and the entropy bonus coefficient. EAPO, with a lower discount factor of $\gent=0.8$ and $\lent=0.95$ surpasses the baseline PPO agent in most of the environments. Additionally, EAPO also outperforms the baseline during the training phase, demonstrating its efficiency in policy optimisation.

We also investigate the impact of the GAE $\lent$ hyperparameter for the entropy advantage estimation. The result shows that $\lent$ does not affect the performance significantly, suggesting that adjusting $\gent$ and $\tau$ would be sufficient for most cases.

\begin{figure*}
    \begin{minipage}{\textwidth}
        \centering
        \includegraphics[width=\textwidth]{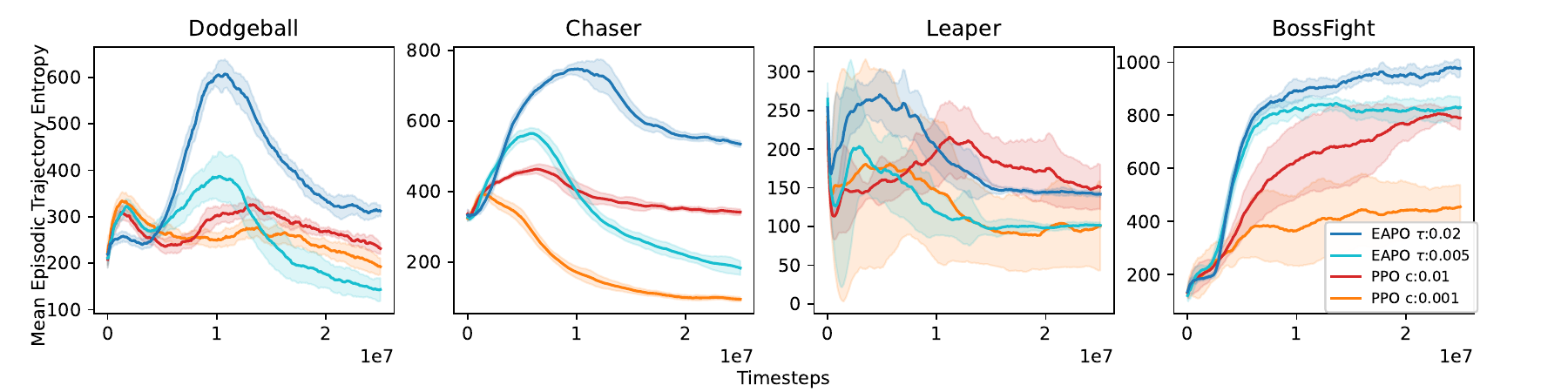}    
    \end{minipage}
     
    \begin{minipage}{\textwidth}
        \centering
        \includegraphics[width=\textwidth]{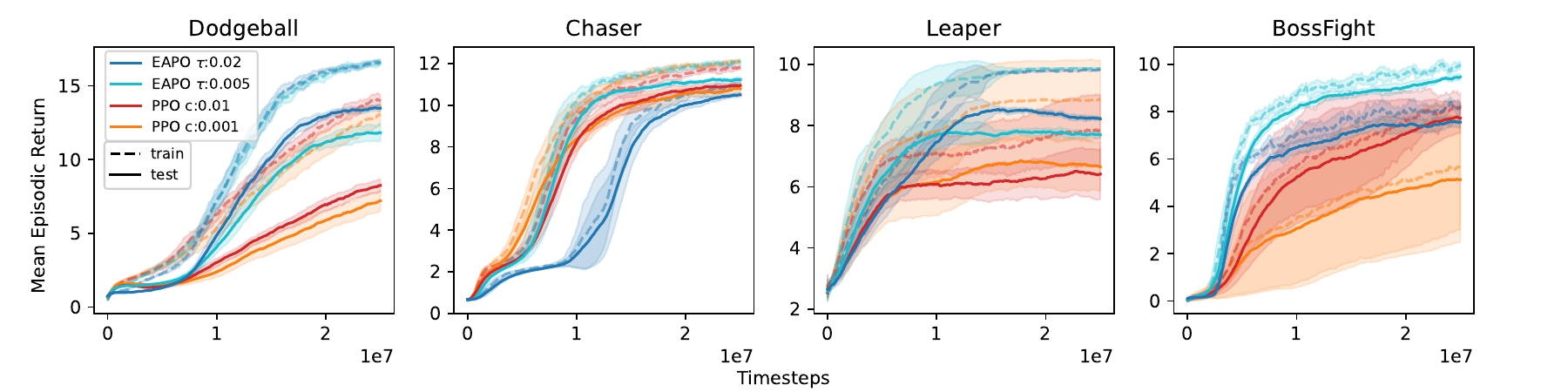}        
    \end{minipage}
    \caption{
        \textbf{Top}: Mean episodic trajectory entropy of EAPO ($\gent=0.8$, $\lent=0.95$) and PPO with entropy cofficients $c\in(0.01, 0.001)$, on a subset of Procgen environments during the test.
        The trajectory entropy of an episode is calculated as the sum of the negative log probability of
        the actions taken in the episode.
        \textbf{Bottom}: Mean episodic return of the selected environments during the test and the training. The higher entropy policy ($\tau=0.02$) outperforms the lower entropy policy ($\tau=0.005$) during the test while achieving matching performance during the training (Dodgeball, Leaper) and exhibits a smaller generalisation gap (Dodgeball, Chaser, Leaper).
    }
    \label{fig:procgen_ent}
\end{figure*}

Figure \ref{fig:procgen} (Right) shows the improved generalisation capability of high entropy policies. The policy trained with higher temperature $\tau$ favours high entropy trajectories (see Figure \ref{fig:procgen_ent}) and performs similar or worse than the one with lower $\tau$ during the training but achieves better during the test. This result is in coherence with the previous study of \citep{eysenbach2021maximum} that a MaxEnt policy is robust to the distributional shift in environments.

\begin{table}
    \caption{
        Mean episodic return at the final timestep of tests on 100 unseen levels on 16 Procgen environments. We report the mean and the 95\% confidence interval from 10 different seeds.
    }
    \label{table:procgen}
    \begin{center}

    \setlength{\tabcolsep}{3pt}
    \begin{tabular}{lllllll}
        \toprule
        \multirow{2}{*}{Env.} & \multicolumn{2}{c}{EAPO $\gent$:0.8 $\lent$:0.95} & \multicolumn{2}{c}{EAPO $\gent$:0.9 $\lent$:0.0} & \multicolumn{2}{c}{PPO} \\
        \cline{2-7} & \multicolumn{1}{c}{\raisebox{-0.5ex}{$\tau$:0.02}} & \multicolumn{1}{c}{\raisebox{-0.5ex}{$\tau$:0.005}} & \multicolumn{1}{c}{\raisebox{-0.5ex}{$\tau$:0.02}} & \multicolumn{1}{c}{\raisebox{-0.5ex}{$\tau$:0.005}}  & \multicolumn{1}{c}{\raisebox{-0.5ex}{\texttt{c}:0.01}} & \multicolumn{1}{c}{\raisebox{-0.5ex}{\texttt{c}:0.001}} \\
        \midrule
        CoinRun & 8.34$\pm$0.24 & 8.31$\pm$0.22 & 7.59$\pm$0.51 & 8.33$\pm$0.36 & 8.13$\pm$0.22 & 7.38$\pm$2.48 \\
        StarPilot & \textbf{54.33$\pm$3.41} & 52.12$\pm$1.95 & 53.28$\pm$2.57 & 54.24$\pm$3.13 & 49.4$\pm$3.72 & 50.57$\pm$2.4 \\
        CaveFlyer & 7.03$\pm$0.32 & 6.48$\pm$0.57 & \textbf{7.17$\pm$0.3} & 6.62$\pm$0.56 & 6.32$\pm$0.7 & 6.3$\pm$0.63 \\
        Dodgeball & \textbf{13.64$\pm$0.68} & 11.59$\pm$0.86 & 13.23$\pm$0.63 & 12.34$\pm$0.85 & 8.51$\pm$0.98 & 7.2$\pm$1.14 \\
        Fruitbot & \textbf{27.28$\pm$0.74} & 26.5$\pm$0.9 & 27.19$\pm$1.14 & 26.57$\pm$1.32 & 25.91$\pm$1.26 & 25.34$\pm$1.18 \\
        Chaser & 10.5$\pm$0.46 & 11.12$\pm$0.3 & 10.52$\pm$0.38 & 11.05$\pm$0.38 & 11.12$\pm$0.45 & 10.64$\pm$0.4 \\
        Miner & \textbf{10.13$\pm$0.38} & 8.9$\pm$0.84 & 10.13$\pm$0.5 & 9.08$\pm$0.78 & 8.6$\pm$0.76 & 8.06$\pm$0.94 \\
        Jumper & 5.7$\pm$0.46 & 5.76$\pm$0.27 & \textbf{5.84$\pm$0.54} & 5.45$\pm$0.48 & 5.21$\pm$0.45 & 5.17$\pm$0.44 \\
        Leaper & \textbf{8.24$\pm$0.52} & 7.75$\pm$0.33 & 7.51$\pm$1.17 & 7.88$\pm$0.67 & 6.54$\pm$0.9 & 6.61$\pm$1.06 \\
        Maze & \textbf{6.5$\pm$0.48} & 5.7$\pm$0.27 & 5.75$\pm$0.52 & 5.56$\pm$0.47 & 5.38$\pm$0.56 & 5.45$\pm$0.31 \\
        BigFish & \textbf{19.89$\pm$2.14} & 17.88$\pm$1.46 & 19.83$\pm$2.4 & 18.73$\pm$2.87 & 12.84$\pm$1.54 & 12.56$\pm$2.86 \\
        Heist & 3.75$\pm$0.66 & 2.97$\pm$0.67 & \textbf{4.22$\pm$0.58} & 3.06$\pm$0.49 & 2.11$\pm$0.54 & 1.68$\pm$0.5 \\
        Climber & 7.43$\pm$0.87 & \textbf{8.0$\pm$0.5} & 6.29$\pm$0.74 & 7.22$\pm$0.52 & 6.78$\pm$0.68 & 6.22$\pm$0.48 \\
        Plunder & 9.0$\pm$0.71 & 7.66$\pm$1.15 & \textbf{9.55$\pm$0.9} & 8.35$\pm$1.38 & 4.16$\pm$0.55 & 3.94$\pm$1.23 \\
        Ninja & 6.49$\pm$0.47 & 6.07$\pm$0.46 & 6.43$\pm$0.41 & 6.21$\pm$0.37 & 6.09$\pm$0.63 & 5.87$\pm$0.58 \\
        BossFight & 7.49$\pm$0.77 & \textbf{9.58$\pm$0.66} & 7.82$\pm$0.71 & 9.38$\pm$0.59 & 7.61$\pm$0.76 & 5.15$\pm$3.32 \\
        \midrule
        Norm. & \textbf{0.54$\pm$0.06} & 0.5$\pm$0.05 & 0.52$\pm$0.07 & 0.51$\pm$0.06 & 0.42$\pm$0.07 & 0.38$\pm$0.11 \\
        \bottomrule
        \end{tabular}

    \end{center}
    \end{table}

\subsection{Discretised continuous control tasks}
\begin{figure*}[t]
    \centering
    \includegraphics[width=0.95\textwidth]{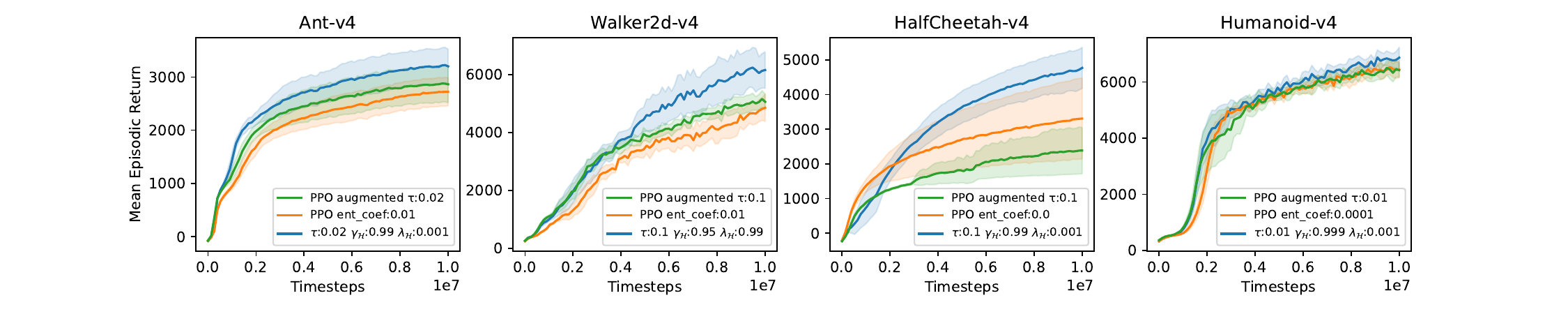}
    \includegraphics[width=0.95\textwidth]{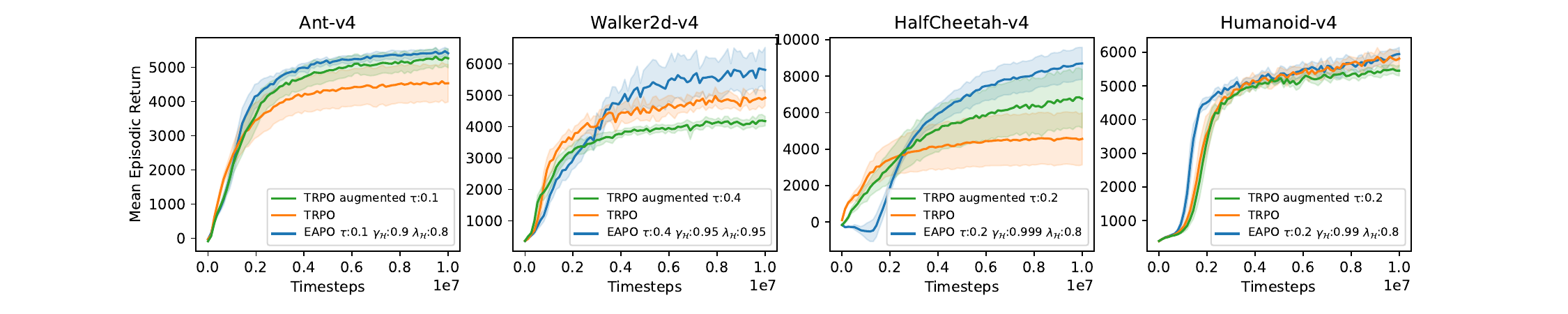}
    \caption{
        Performance comparison on 4 MuJoCo tasks. We measured the mean episodic return of the stochastic policy periodically over 100 episodes during the training. Results are averaged from 10 random seeds, and the shaded area indicates the 95\% confidence interval.  \textbf{Top}: EAPO-PPO. We compare EAPO to the PPO agent with the best-performing entropy coefficient, and with the entropy reward augmented PPO. \textbf{Bottom}: EAPO-TRPO. We also compare with the entropy reward augmented TRPO.
    }
    \label{mujoco_best}
\end{figure*}

We measure the performance of EAPO extending PPO (EAPO-PPO) and TRPO (EAPO-TRPO) on continuing control tasks in 4 MuJoCo environments, comparing them against their corresponding baselines. For the PPO baselines, we searched for the best entropy coeffcient within the set $c\in(0.0001, 0.001, 0.01)$. Additionally, we tested the PPO and TRPO agent with the reward augmented by the entropy reward $-\tau\lpt$ to evaluate the impact of separating the MaxEnt objective. Note that the entropy reward-augmented baseline is effectively regarded as EAPO with $\gent = \gamma$ and $\lent = \lambda$, but without the entropy critic. We discretise the continuous action space using the method proposed by \citet{tang2020discretizing}. Experiments using continuous action space are provided in Appendix \ref{app:ae}. The training curves are presented in Figure \ref{mujoco_best}.

The result shows that by adjusting $\gent$ and $\lent$, we can configure EAPO to outperform or match the conventional entropy regularisation method throughout all environments. We found that the best-performing values of $\gent$ and $\lent$ vary depending on the characteristics of the environment, similar to their value counterparts $\gamma$ and $\lambda$, respectively. Although EAPO demonstrates more stable performance compared to the entropy bonus, this relatively modest performance gain suggests that EAPO may be less efficient for continuing tasks.

Figure \ref{mujoco_best} also demonstrates that the adjustability adopted by EAPO improves the naive implementation of the MaxEnt policy that augments the entropy reward. We also provide ablation experiments on $\gent$ and $\lent$ using PPO-based EAPO in Appendix \ref{app:ae}.

\subsection{\texttt{MiniGrid-DoorKey-8x8} environment}
\begin{figure*}[t]
    \begin{minipage}{0.5\textwidth}
        \centering
        \includegraphics[width=0.9\textwidth]{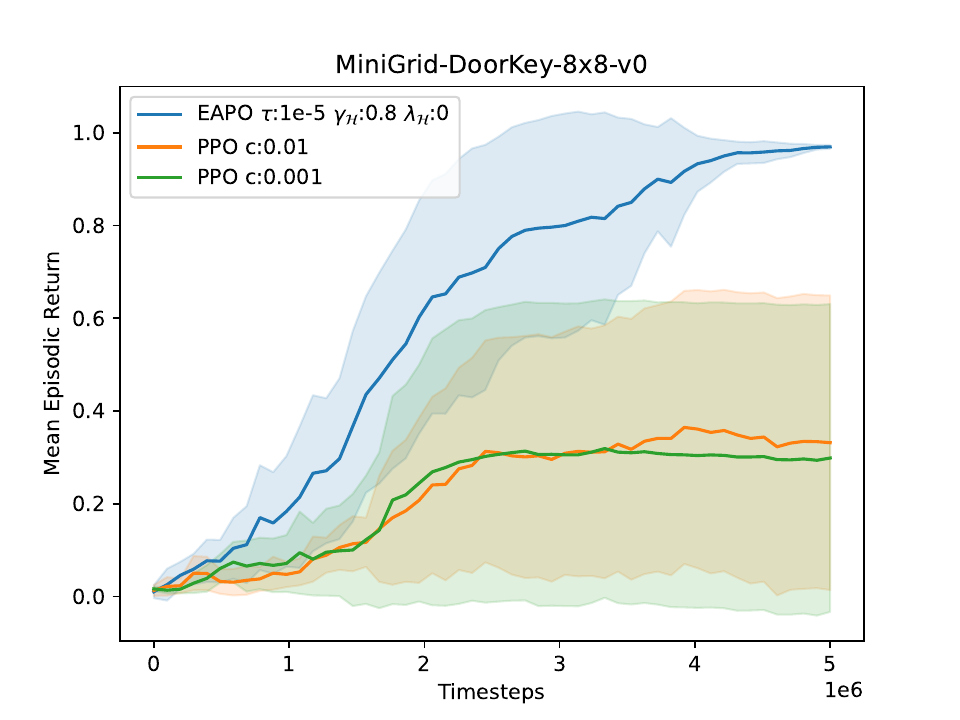}    
    \end{minipage}
    \hfill
    \begin{minipage}{0.5\textwidth}
        \centering
        \includegraphics[width=0.9\textwidth]{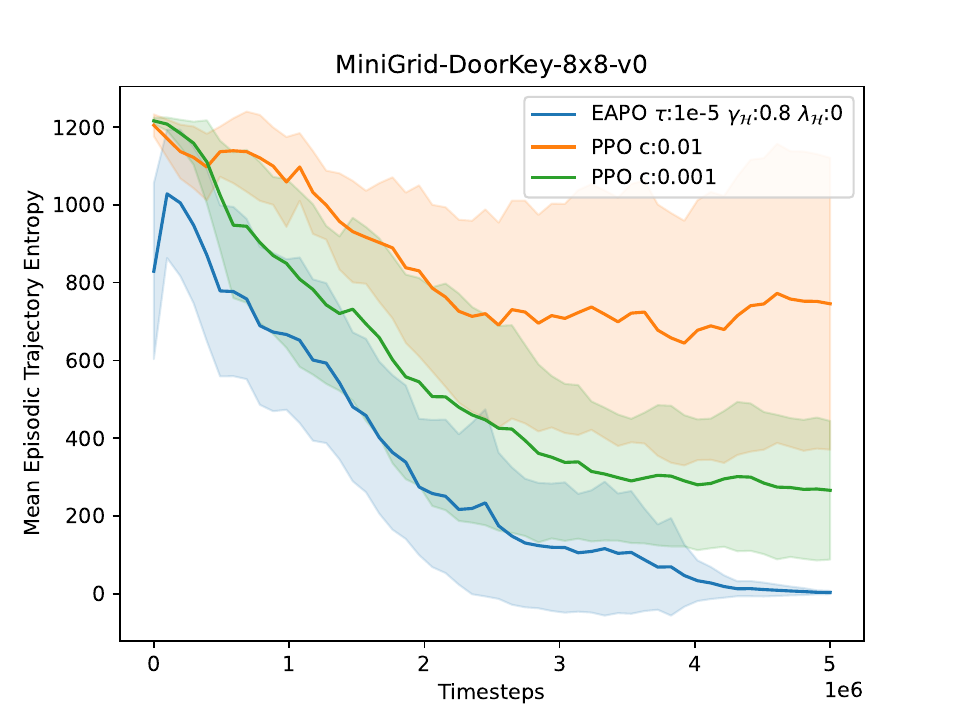}        
    \end{minipage}
    \caption{
        The return and trajectory entropy comparison results of EAPO with $\tau=1e-5$, $\gent=0.8$ and $\lent=0$ and PPO with entropy coefficent $0.01$ and $0.001$. Results are averaged from 10 random seeds, and the shaded area indicates the 95\% confidence interval. 
    }
    \label{dk}
\end{figure*}

Finally, we evaluate the exploration performance of PPO-based EAPO on the \texttt{MiniGrid-DoorKey-8x8} environment \citep{MinigridMiniworld23}. Figure \ref{dk} shows that EAPO, with the given hyperparameters, can solve this hard exploration task within 5,000,000 frames for all 10 seeds, whereas the baseline PPO only achieves the goal for 3 seeds. However, unlike other tasks presented in this work, this task was highly sensitive to hyperparameters. As noted by \citep{mei2020global}, entropy regularization may not effectively mitigate epistemic uncertainty. Our results show inconclusive evidence for better exploration using MaxEnt RL in this context.

\section{Conclusion}
We have introduced EAPO, a model-free on-policy actor-critic algorithm based on the maximum entropy reinforcement learning framework.
Our approach shows that a straightforward extension of existing mechanisms for standard value learning in on-policy actor-critic algorithms to the trajectory entropy objective can facilitate the practical implementation of MaxEnt RL.
Through empirical evaluations, our method has been shown to replace the conventional entropy regularisation method and that a more principled entropy maximisation method enhances generalisation.
While this paper focuses on PPO and TRPO, one can seamlessly adapt EAPO to other advantage actor-critic algorithms such as A3C \citep{mnih2016asynchronous}. This adaptability lays the groundwork for deeper investigations into the interactions between the MaxEnt RL framework and various components of reinforcement learning algorithms.
We anticipate that the inherent simplicity of on-policy algorithms and EAPO will encourage broader applications of the MaxEnt RL algorithm to promising areas like competitive reinforcement learning and robust reinforcement learning.

\clearpage

\clearpage
\appendix

\section{Proof of the soft policy gradient theorem} \label{app:spgt}
We begin with the proof of \citet{shi2019soft}. Let us denote the discounted state distributions induced by the policy $\pi$ as $\rho^\pi(s)$ for the discount factor $\gamma_V$, and as $\rho^\pi_\ent(s)$ for the discount factor $\gent$, respectively. Then,
\begin{align*}
    \nabla_\theta J(\pi_\theta) 
    &= \expect{\substack{s_0\sim \rho, \\ a_t\sim\pi, \\ s_{t+1}=\mathcal{T}(s_t,a_t)} }{(\tilde{Q}^\pi(s_t, a_t) - \tau\log\pi(a_t|s_t) - 1)\nabla_\theta\log\pi(a_t|s_t)} \\
    &= \sum_{s}\left[\rho^\pi(s)[\nabla_\theta\sum_a\pi(s|a)[Q^\pi(s, a) - V^\pi(s)]] \right. \\ 
    & \left. \qquad+ \tau\rho^\pi_\ent(s)[\nabla_\theta\sum_a\pi(s|a)(\qent(s,a) -\log\pi(a|s))]\right] \\
    &= \expect{s\sim\rho^\pi,a_t\sim\pi}{\nabla_\theta\log\pi_\theta(a_t|s_t)A^\pi(s_t,a_t)} + \tau\expect{s_t\sim\rho^\pi_\ent,a_t\sim\pi}{\nabla_\theta\log\pi_\theta(a_t|s_t)\aent(s_t,a_t)} \\
    &= \expect{\substack{s_0\sim \rho, \\ a_t\sim\pi, \\ s_{t+1}=\mathcal{T}(s_t,a_t)} }{\nabla_\theta\log\pi_\theta(a_t|s_t)\gamma_V^t A^\pi(s_t,a_t)} \\ 
    &\qquad+ \tau\expect{\substack{s_0\sim \rho, \\ a_t\sim\pi, \\ s_{t+1}=\mathcal{T}(s_t,a_t)} }{\nabla_\theta\log\pi_\theta(a_t|s_t)\gent^t \aent(s_t,a_t)} \\
    &= \expect{\substack{s_0\sim \rho, \\ a_t\sim\pi, \\ s_{t+1}= \mathcal{T}(s_t,a_t)} }{(\gamma_V^t A(s_t,a_t)+\gent^t \tau\aent(s_t,a_t))\nabla_\theta\log\pi_\theta(a_t|s_t)}.
    \tag*{$\square$}
\end{align*}

\section{Hyperparameters and implementation details} \label{app:hps}
\subsection{Details of the \texttt{MiniGrid-Empty-8x8-v0} example}\label{app:example-empty}
The action space of \texttt{MiniGrid-Empty-8x8-v0} \citep{MinigridMiniworld23} consists of 7 discrete actions: 2 actions for turning left or right, 1 action for moving forward, and 4 no-op actions. The reward, calculated as  $1 - 0.9t/T$ is given only when the agent reaches the goal state, where $t$ is the number of steps taken and $T$ is the maximum number of steps allowed. For $T$, the default value (256) is used. We modified the turning actions so that they also move the agent forward to the corresponding directions (i.e., no extra step is required to maneuver), enabling multiple optimal trajectories. This modification reduces the optimal number of steps from 11 to 10.  The observation used is the full $8\times 8 \times 3$ image without partial observability. We employ a simple CNN architecture from \citep{rl-starter-files} as a shared feature extractor. The state visitation plots are generated from 100 rollouts using policies trained over 4M frames. The mean trajectory entropy and the mean steps are averaged from 10 different random seeds. We select the representative heatmap for each setup from the run with the trajectory entropy closest to the average trajectory entropy across all seeds. The normalised state visitation frequency is calculated by dividing the number of visits to each state divided by the total number of state visits in the trajectories.

\subsection{Hyperparameters}
We use the default values in stable-baseline3 \citep{raffin2019stable} and envpool \citep{weng2022envpool} libraries for the settings not specified in the table \ref{table:ppo hp}. EAPO-specific hyperparameters are reported in Table \ref{table:eapo_hp}. The parameters for the MuJoCo tasks are found in a coarse hyperparameter search.
\setlength{\tabcolsep}{3pt}
\begin{table}[h]
    \caption{Common hyperparameters.}
    \label{table:ppo hp}
    \begin{center}
        \begin{tabular}{llllll}
            \toprule
            Parameter &                                             MuJoCo (PPO) & MuJoCo (TRPO) &                                            Procgen  & Empty & DoorKey\\
            \midrule
            Timesteps               &                                           10M &   10M&                                        25M & 4M & 5M\\
            Num. Envs                        &                                                 64 & 64 &                                                64 & 16 & 64 \\
            Num. Steps                       &                                                128 &     64 &                                           256  & 128 & 128\\
            Learning Rate                 &                                            \num{5e-4} &  \num{5e-4} &                                            \num{5e-4} & \num{5e-4} & \num{1e-3} \\
            Batch Size                    &                                               1024 &                 1024 (critic)&                              2048 & 1024 & 1024 \\
            Epochs                      &                                                 4 &    4 (critic)   &                                           3 & 4 & 4 \\
            Discount Factor                          &                                               0.99 &  0.99  &                                          0.995 &0.99 & 0.995 \\
            GAE $\lambda$                    &                                               0.95 &    0.95              &                              0.8 & 0.95 & 0.995 \\
            Clip Range $\epsilon$                    &                                               0.2 &0.2 &                                               0.1 & 0.2 & 0.2 \\
            Max Grad. Norm.                &                                                3.5 & 3.5&                                                0.5 & 0.5 & 0.5 \\
            Adv. Norm. &                                               True & True &                                              True & True & True \\
            Obs. Norm.                      &                                               True & True &                                              False & False & False \\
            Rew. Norm.                   &                                              False &  False &                                             False & False & False \\
            PopArt $\beta$                  &                                               0.03 & 0.03 &                                              0.03 & 0.03 & 0.03 \\
            Value Loss Coeff. & 0.25 & 0.25 & 0.5 & 0.5 & 0.5 \\
            Entropy Coeff. & (1e-2, 1e-3, 1e-4) & (1e-2, 1e-3, 1e-4) & (1e-2, 1e-3) & None & [1e-5, 1e-4] \\
            Target KL & None & 0.07 &  None & None & None\\
            \bottomrule
            \end{tabular}
    \end{center}
\end{table}
\setlength{\tabcolsep}{3pt}
\begin{table}[h]
    \caption{EAPO specific hyperparameters.}
    \label{table:eapo_hp}
    \begin{center}
        \begin{tabular}{lllllll}
            \toprule
            Parameter                & Procgen                      & Empty                         & DoorKey                &             & MuJoCo (PPO) & MuJoCo (TRPO)   \\
            \midrule
            \multirow{4}{*}{$\gent$} & \multirow{4}{*}{(0.8, 0.9)}  & \multirow{4}{*}{(0.9, 0.99)}  & \multirow{4}{*}{0.8}   & Ant         & 0.99         & 0.9             \\
                                     &                              &                               &                        & Walker2d    & 0.95         & 0.95            \\
                                     &                              &                               &                        & HalfCheetah & 0.99         & 0.999           \\
                                     &                              &                               &                        & Humanoid    & 0.999        & 0.99            \\
            \hline                                     
            \multirow{4}{*}{$\lent$} & \multirow{4}{*}{(0.95, 0.0)} & \multirow{4}{*}{0.0}          & \multirow{4}{*}{0.0}   & Ant         & 0.001        & 0.8             \\
                                     &                              &                               &                        & Walker2d    & 0.99         & 0.95            \\
                                     &                              &                               &                        & HalfCheetah & 0.001        & 0.8             \\
                                     &                              &                               &                        & Humanoid    & 0.001        & 0.8             \\
            \hline
            \multirow{4}{*}{$\tau$}  & \multirow{4}{*}{(0.2, 0.005)}& \multirow{4}{*}{0.0}          & (0.001,                & Ant         & 0.02         & 0.1             \\
                                     &                              &                               & 0.002,                 & Walker2d    & 0.1          & 0.4             \\
                                     &                              &                               & 0.003)                 & HalfCheetah & 0.1          & 0.2             \\
                                     &                              &                               &                        & Humanoid    & 0.01         & 0.2             \\
            \hline
            $c_2$                    & 0.5                          & 1.0                           & 1.0                    &             & 1.0          & 1.0             \\
            \bottomrule
        \end{tabular}
    \end{center}
\end{table}

\subsection{Network architecture}
For discretised MuJoCo tasks, we used simple $\tanh$ networks for the policy and the critics with hidden layers of depth [64, 64] and [128, 128], respectively. We implemented the entropy critic as the independent output layer that shares the hidden layers with the value network. For the continuous MuJoCo tasks we use [256, 256] and [512, 512] for policy and critic networks and the state- and action-dependent $\sigma$ output for the Gaussian policy with the Softplus output layer.

In Procgen benchmark experiments, we adopt the same IMPALA CNN architecture used in \citet{cobbe2020leveraging}. The entropy critic is again a single final output layer that shares the CNN feature extractor.

\section{Additional experiments}\label{app:ae}

\begin{figure*}
    \centering
    \includegraphics[width=0.95\textwidth]{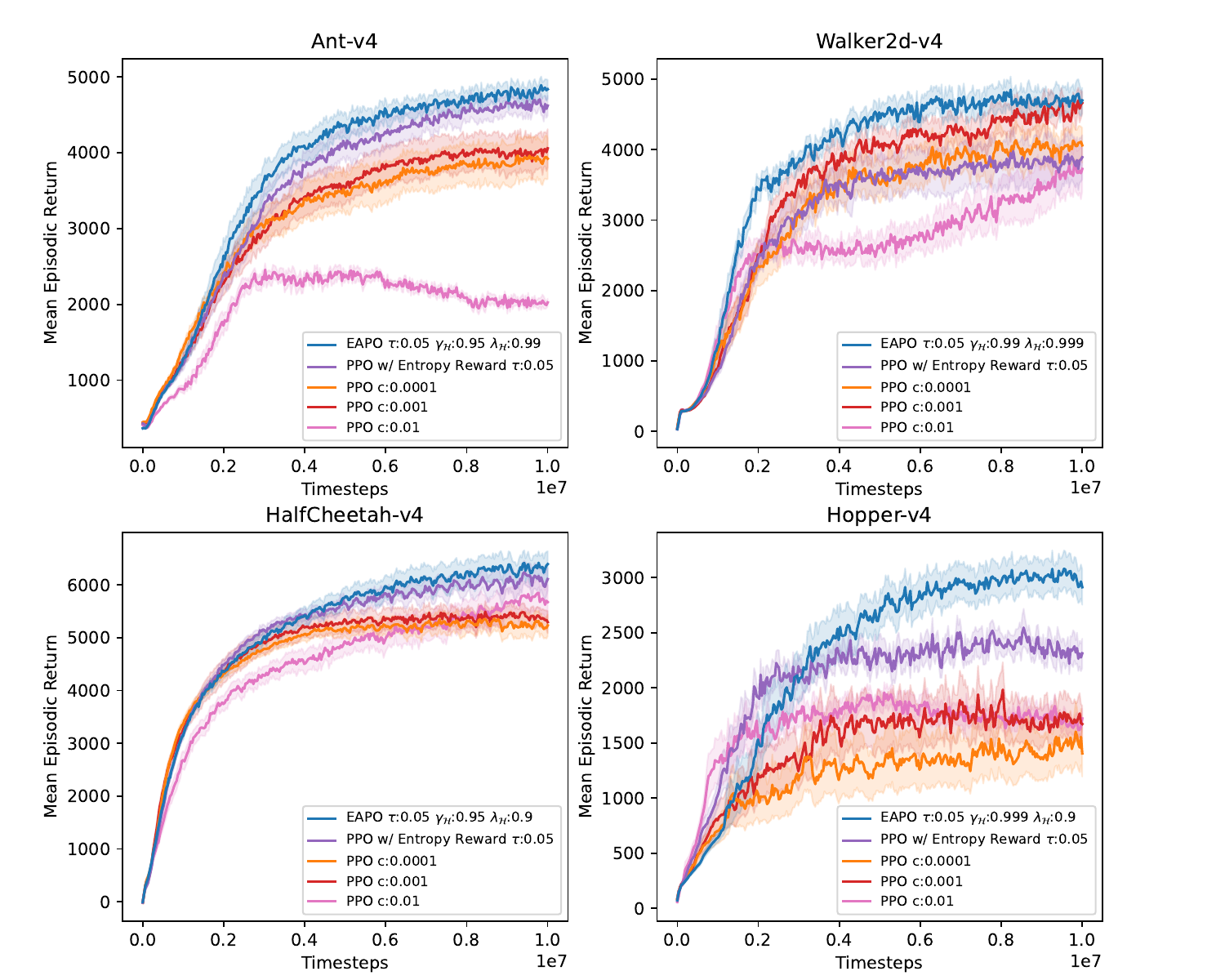}
    \caption{
        Performance comparison on 4 MuJoCo continuous locomotion tasks.
    }
    \label{fig:mujoco_cont}
\end{figure*}

\begin{figure*}
    \centering
    \includegraphics[width=0.99\textwidth]{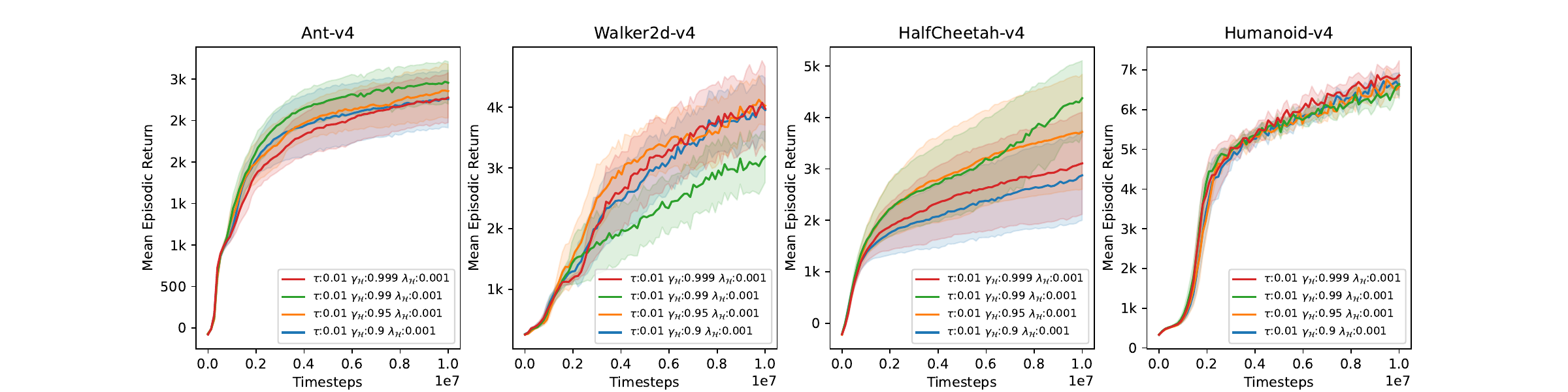}
    \includegraphics[width=0.99\textwidth]{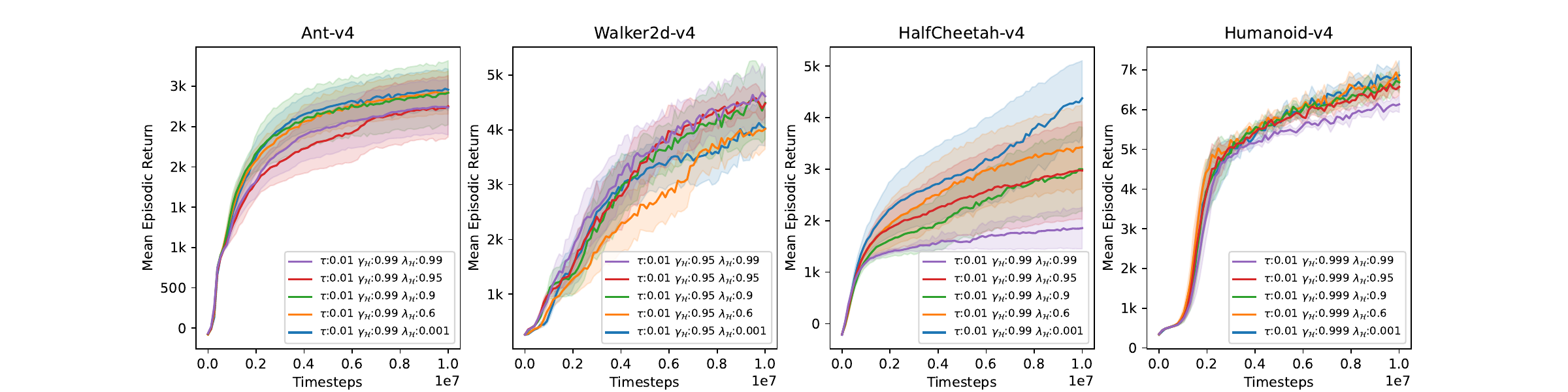}
    \includegraphics[width=0.99\textwidth]{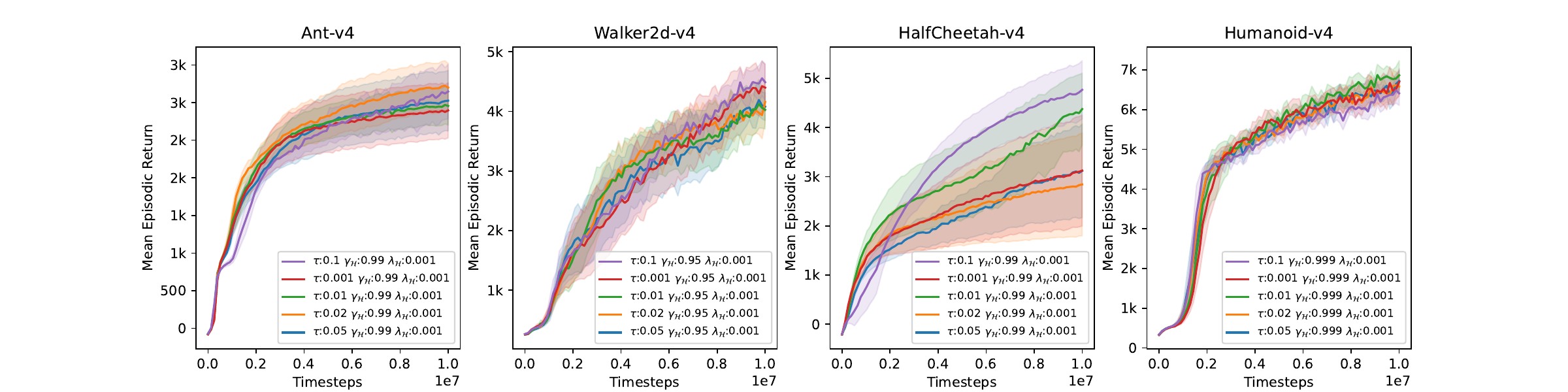}
    \caption{
        The ablation results on $\gent$, $\lent$, and $\tau$ for PPO-based EAPO.
    }
    \label{fig:mujoco_abl}
\end{figure*}

\end{document}